\documentclass[runningheads]{llncs}

\usepackage{eccv}

\usepackage{eccvabbrv}

\usepackage{graphicx}
\usepackage{booktabs}

\usepackage[accsupp]{axessibility}  % Improves PDF readability for those with disabilities.

\usepackage{amsmath,amsfonts,bm}

\def\eqref#1{equation~\ref{#1}}
\def\1{\bm{1}}

\DeclareMathAlphabet{\mathsfit}{\encodingdefault}{\sfdefault}{m}{sl}
\SetMathAlphabet{\mathsfit}{bold}{\encodingdefault}{\sfdefault}{bx}{n}

\DeclareMathOperator*{\argmin}{arg\,min}

\newcommand{\parag}[1]{\vspace{-3mm}\paragraph{#1}}

\newcommand{\sssr}[0]{$\mathit{SVR}econ$}
\usepackage{url}
\usepackage{bbm}
\usepackage{pifont}
\usepackage{algorithm}
\usepackage{algpseudocode}
\usepackage{graphicx}
\usepackage{enumitem}
\usepackage{wrapfig}
\usepackage{multirow}

\usepackage[square,numbers]{natbib}

\usepackage[breaklinks,colorlinks,citecolor=eccvblue]{hyperref}

\usepackage{orcidlink}

\begin{document}

\title{Generalizable Neural Reconstruction of High-Fidelity Surfaces via Sparse Volumetric Representations}

\titlerunning{Generalizable Neural Reconstruction via Sparse Volumetric Representations}

\author{Aoxiang Fan\inst{1}\orcidlink{0000-0002-2877-9795} \and
Corentin Dumery\inst{1}\orcidlink{0000-0001-5314-7979} \and
Nicolas Talabot\inst{1}\orcidlink{0000-0003-0952-5906} \and
Ming Xu\inst{1}\orcidlink{0000-0002-6478-0582} \and
Hieu Le\inst{2}\orcidlink{0000-0001-7855-2778} \and
Pascal Fua\inst{1}\orcidlink{0000-0002-6702-9970}}

\authorrunning{A. Fan et al.}

\institute{
CVLab, École Polytechnique Fédérale de Lausanne (EPFL), Switzerland\\
\email{\{aoxiang.fan, corentin.dumery, nicolas.talabot, mingda.xu, pascal.fua\}@epfl.ch}
\and
University of North Carolina at Charlotte, USA\\
\email{hle40@charlotte.edu}
}

\maketitle

% !TEX root = ../main.tex
% !TEX spellcheck = en-US

\begin{abstract}

Neural implicit representations have recently achieved impressive results in novel view synthesis and multi-view 3D reconstruction, yet both NeRF- and Gaussian Splatting-based methods require per-scene optimization, which makes them inefficient. Generalizable Neural Surface Reconstruction (GNSR) methods have been proposed to remove this need by learning feature representations directly predicted from input images. However, their typical reliance on dense feature volumes severely limits achievable resolution and fidelity due to prohibitive memory costs. We introduce Sparse Volumetric Reconstruction (SVRecon), a new GNSR framework that unlocks high-resolution, memory-efficient reconstruction through learned occupancy-driven sparsity, in a more effective way than earlier approaches to introducing sparsity in GNSRs. Our approach uses a nested two-stage architecture: (1) an occupancy prediction network that identifies surface-containing voxels, and (2) a high-resolution sparse volume rendering framework defined only within these occupied regions, together with specialized sparsified algorithms for ray sampling, feature aggregation, and querying. This design enables fine-grained surface reconstruction while avoiding the heavy memory footprint of dense grids. SVRecon operates at resolutions up to $512^3$ on standard 32GB hardware—substantially higher than prior generalizable methods—and delivers smoother and more precise reconstructions across diverse datasets, particularly in sparse-view settings.

\end{abstract}    
% !TEX root = ../main.tex
% !TEX spellcheck = en-US

\section{Introduction}
\label{sec:intro}

The emergence of neural implicit representation techniques, starting with the seminal Neural Radiance Fields (NeRF)~\citep{Mildenhall20} and continuing with Neural Implicit Surfaces (NeuS)~\citep{Yariv20,Wang21f,Yariv21,Li23c}, has greatly boosted novel view synthesis and multi-view geometry reconstruction. These implicit surface reconstruction algorithms take posed images as input, and optimize a Signed Distance Function (SDF) to minimize a volume rendering loss, which yields accurate 3D reconstructions given enough views. A major drawback, however, is that the optimization must be performed from scratch for every new set of views, which makes these approaches computationally demanding. Furthermore, accuracy tends to decline when only a few views are available. Newer 3D Gaussian Splatting-based methods~\citep{kerbl23} are subject to the same limitations~\citep{Huang24b,Yu24}.

Generalizable approaches to Neural Surface Reconstruction (GNSRs)~\citep{Liang24,Ren23a,Xu23f,Long22b,Na24,Younes24} are designed to remove these restrictions. These methods are pre-trained on a large number of scenes. As a result, the rendering of a novel scene simply becomes a test-time inference procedure, without any optimization required. Volume rendering and scene reconstruction are performed by relying on a 3D scene representation produced by a neural network that takes as input the images. This representation is typically in the form of a dense volume of high-dimensional features. However, even though they are much faster, these methods have yet to achieve their full potential because the necessary volumetric feature representation is extremely memory-intensive. This drastically constrains their operational resolution, and adversely impacts reconstruction quality and detail preservation.

Sparse scene representation is a natural answer to this problem. Since surfaces are mathematically of measure zero, they occupy only a small fraction of voxels in a discretized 3D space, which may be leveraged for high-resolution, memory-efficient reconstruction. However, this brings significant challenges. 
Existing approaches to sparsification~\citep{Long22b,Peng24b} require a complex and cumbersome pre-rendering process to infer sparsity followed by using densified feature volumes in the final rendering stage, which negates many of the benefits of sparsification. In contrast, our occupancy-prediction network offers fine-scale sparsity prediction in a simple feed-forward process. Moreover, our single-scale occupancy field can be combined with our specialized algorithms for efficient sparse volume rendering. As a result, our model can operate at a $512^3$ resolution, which existing methods cannot do as shown in Tab.~\ref{tab:resolution}.

% !TEX root = ../main.tex
% !TEX spellcheck = en-US

\begin{figure*}[!t]
    \centering
    \includegraphics[width=0.95\textwidth]{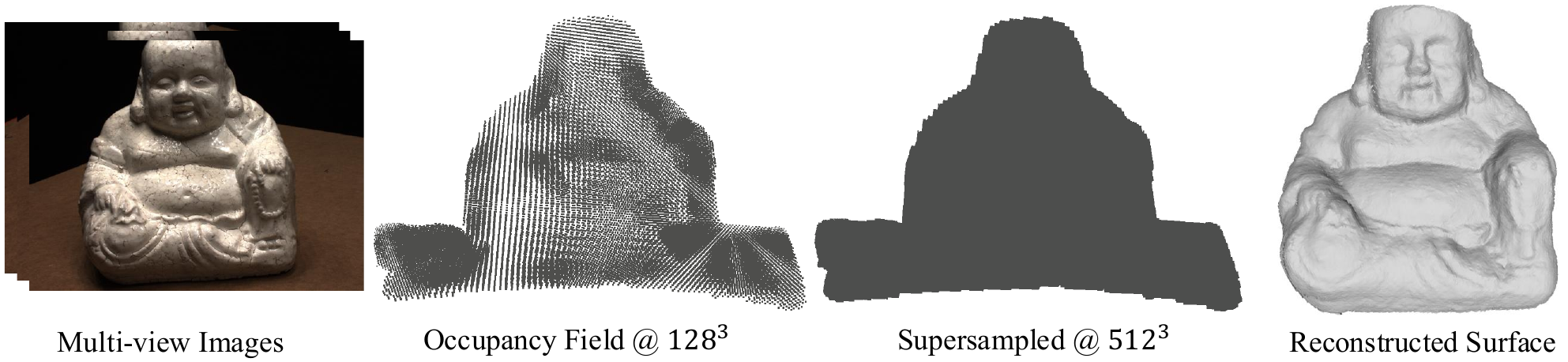}
    \caption{\textbf{Two-Stage Approach}. From multiple images, we predict occupancy field that is then supersampled to build a sparse scene representation. This lets us reconstruct high-fidelity surfaces by operating at previously unattainable $512^3$ resolution. }
    \label{fig:illustration}
\end{figure*}

In general terms, we propose a novel and effective way of bringing sparsity into GNSRs with occupancy predictions. Specifically, we introduce a nested two-stage architecture that enables our GNSR to operate at significantly higher resolutions than previous ones, while maintaining both generalizability across scenes and the ability to handle cases where only a limited number of views are available.  As illustrated by Fig.~\ref{fig:illustration}, the two stages perform the following tasks:
\begin{enumerate}[itemsep=1pt, topsep=0pt]

  \item {\bf Computing occupancy at a lower resolution.} Voxels from a coarser grid are labeled as containing a surface element or not, by training a network using posed images as input. We implement this network to significantly sparsify the 3D volume while achieving near-perfect recall.

  \item {\bf Volume rendering with sparse representations at a high resolution.} High-dimensional features are constructed at a higher-resolution but only within the occupied voxels, where volume rendering can be performed. This allows for fine-grained neural surface reconstruction while keeping memory and computational costs at a minimum.

\end{enumerate}
%
% !TEX root = ../main.tex
% !TEX spellcheck = en-US

\begin{table}[t!]
    \centering
    \resizebox{0.45\linewidth}{!}{
    \begin{tabular}{c|c|c|c|c}
    \hline
    Methods & $128^3$ & $192^3$ & $256^3$ & $512^3$ \\ \hline
    SparseNeuS~\cite{Long22b} & \ding{51} & \ding{51} & \ding{55} & \ding{55} \\
    VolRecon~\cite{Ren23a} & \ding{51} & \ding{55} & \ding{55} & \ding{55} \\
    ReTR~\cite{Liang24} & \ding{51} & \ding{55} & \ding{55} & \ding{55} \\
    Ours (\sssr{}) & \ding{51} & \ding{51} & \ding{51} & \ding{51} \\
    \hline
    \end{tabular}
    }
    \vspace{0.2cm}
    \caption{{\bf Resolutions handled by GNSR algorithms at training time}. Crosses indicate that the memory requirements were too large on a 32GB NVIDIA Tesla V100 GPU, with the same setting of batch size $1$ and $1024$ sampled rays for all methods.}
    \label{tab:resolution}
\end{table}

\begin{figure*}[t]
    \centering
    \includegraphics[width=0.99\textwidth]{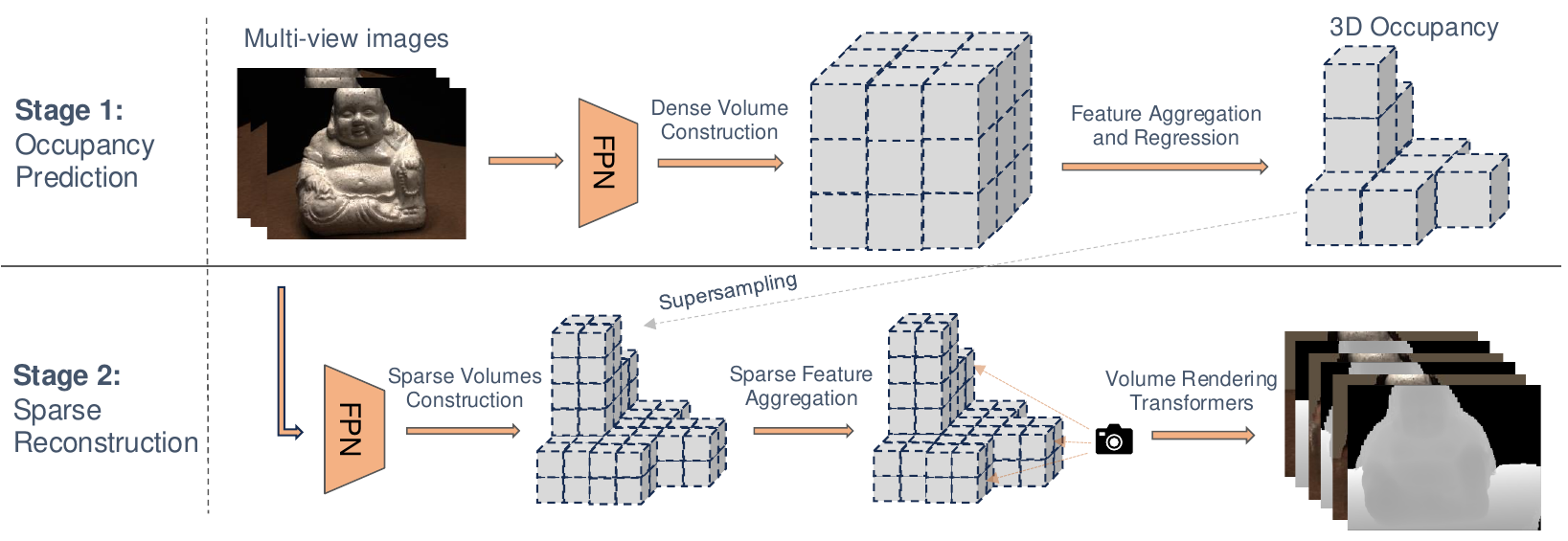}
    \caption{\textbf{\sssr{} pipeline.} First, we build a low-resolution dense 3D feature volume from multi-view features and predict 3D occupancies. Second, we construct high-resolution sparse feature volumes where the predicted occupancy warrants it. Finally, volume rendering Transformers are used to aggregate per-ray sampled features and infer color and depth, enabling scene reconstruction. }
    \label{fig:pipeline}
\end{figure*}

Fig.~\ref{fig:pipeline} summarizes our \textit{Sparse Volumetric Reconstruction} approach, which we dub \sssr{}. By scaling up the operating resolution, our main contribution firstly includes demonstrating the significance of 3D sampling resolutions for higher-fidelity reconstruction. In addition, our new method requires rethinking and reformulating the algorithms used by existing methods, which are designed for handling dense representations. Thus, our contribution also includes a dedicated framework for sparse volume rendering, supporting sparse operations including ray sampling, feature aggregation, and querying. These technical innovations are key to realizing the theoretical memory advantages of our approach while maintaining computational efficiency. 

Testing on public datasets demonstrates that our sparse-representation method is highly effective, enabling us to operate at a high resolution of $512^3$ on standard hardware with 32GB of VRAM. Consequently, our method produces the smoothest and most precise surfaces of all state-of-the-art methods.
% !TEX root = ../main.tex
% !TEX spellcheck = en-US

\section{Related Works}
\label{sec:related_works}

\paragraph{Multi-View Stereo.} Multi-view stereo (MVS) has long been known to be effective for 3D reconstruction~\citep{Fua97a,Hartley00,Shum00b,Seitz06,Furukawa09b,Lhuillier05,Kostrikov14,Kutulakos00}. Modern multi-view stereo methods can be categorized into depth-map ones~\citep{Yao18,Yao19,Gu20b,Wang22f,Ding22,Cao24} and volumetric ones~\citep{Ji17b,Ji20,Kar17}. Depth-map methods rely on view-dependent frustums as an indirect model of 3D space, which limits the reconstruction accuracy without providing novel view rendering capabilities. Volumetric methods are directly related to our method, as they also operate on a volumetric 3D space. Comparatively, our scene representation and volume rendering formulation make it possible to recover finer details.

% !TEX root = ../main.tex
% !TEX spellcheck = en-US

\parag{Neural Scene Reconstruction.} The success of NeRFs~\citep{Mildenhall20} has prompted researchers to examine the use of Neural Implicit Representations for surface reconstruction as an alternative to MVS. This line of work includes IDR~\citep{Yariv20}, VolSDF~\citep{Yariv21}, NeuS~\citep{Wang21f}, and Neuralangelo~\citep{Li23c}, which are able to recover fine geometry given dense input views. More recently, SparseCraft~\citep{Younes24} proposes to use stereopsis cues regularization to enhance surface reconstruction qualities under sparse input views. However, all these methods are not generalizable and need to be trained per scene.

\parag{Generalizable Surface Reconstruction.} Many generalizable approaches have been proposed to dispense with intensive per-scene optimization. These methods include volumetric methods such as VolRecon~\citep{Ren23a}, ReTR~\citep{Liang24}; and frustum-based 3D representations~\citep{Na24,Xu23f}. However, all these methods have to balance reconstruction quality against storage overhead due to the cost of their global scene representations, which is what our method improves upon. Some of these methods~\citep{Long22b,Peng24b} incorporate some sparsity handling. However, they depend on a cumbersome pre-rendering process, and still rely on densified feature volumes in the fine-rendering stage, which mostly negates the benefits of sparsification. As a result, our more streamlined approach can achieve a much higher resolution, as shown by Tab.~\ref{tab:resolution}, which lists the admissible resolutions the various algorithms can handle. 

\parag{Gaussian Splatting.} Gaussian splatting~\citep{kerbl23} has emerged as a powerful alternative to NeRF methods for novel view synthesis. The same trend is at work for 3D reconstruction, with the advent of methods such as 2D Gaussian Splatting~\citep{Huang24b} and Gaussian Opacity Fields~\citep{Yu24}. There are also recent attempts at making Gaussian Splatting generalizable~\citep{Zhang25d,Yang25}. However, they typically struggle to reconstruct smooth surfaces due to the inherent difficulty of modeling surfaces with Gaussians.

% !TEX root = ../main.tex
% !TEX spellcheck = en-US

\section{Background: Generalizable NeRFs}

We now briefly review the general formulation of  the original and generalizable NeRFs. In both cases, the goal is to reconstruct a scene captured by a set of posed images $\mathcal{I} = \{(\mathbf{I}_i, \mathbf{P}_i)\}_{i=1}^M$, where $\mathbf{I}_i \in \mathbb{R}^{H \times W \times 3}$ and $\mathbf{P}_i \in \mathbb{R}^{3 \times 4}$ denote the projection matrix associated with the $i$-th view.

Given $\mathcal{I}$ as input, the aim is to learn a scene representation function $F$ and a rendering function $R$ 
\begin{align}
    F: & (\mathbf{x}, \mathbf{r}) \rightarrow \mathbf{f}, \label{eq:representation} \\
    R: & \{(\mathbf{x}_i, \mathbf{f}_i)\}_{i=1}^N \rightarrow (\mathbf{c}_i, d_i),
\end{align}
where for the inputs, $\mathbf{x}_i \in \mathbb{R}^3$ are 3D locations (typically sampled along a ray extending from the camera frustum) and $\mathbf{r} \in \mathbb{S}^2$ is a viewing direction, and for the outputs, $\mathbf{f} \in \mathbb{R}^C$ is a feature vector, $\mathbf{c}_i\in\mathbb{R}^3$ is a color and $d_i \in \mathbb{R}^+$ is (optional) depth.

In the original NeRF, $\mathbf{f}$ directly encodes an emitted color and a volume density, $R$ is a handcrafted volume rendering function. The scene representation function $F$ is implemented by a network with weights $\lambda$. Learning them requires per-scene minimization of the photometric loss
\begin{align}
\label{eq:nerf_opt}
\argmin_{\lambda} \sum_{i=1}^M \sum_{\mathbf{p} \in \mathbf{I}_i} L(\mathbf{c}_{i}, \hat{c}(I_i, \mathbf{p})) \; ,
\end{align}
where $L$ is usually taken to be square norm of the difference between the arguments, $\mathbf{p}$ are pixel coordinates and $\hat{c}(I, \mathbf{p})$ yields the color at $\mathbf{p}$ in image $I$.

In a generalizable NeRF, the feature vector $\mathbf{f}$ is high-dimensional and does not have physical meaning. The handcrafted rendering function $R$ is replaced by one implemented by a network with weights $\phi$,  while $F$ uses a different architecture that allows for generalization and has weights $\theta$. The weights $\theta$ and $\phi$ are learned by training on a large corpus of scenes $\mathcal{D} = \{(\mathcal{I}_i, \mathbf{D}_i)\}_{i=1}^{J}$, where $\mathbf{D}_i\in\mathbb{R}^{H\times W}$ denotes the ground-truth depth map corresponding to image $\mathcal{I}_i$. We minimize
% $\mathsf{G}_\lambda$, so that  $\mathsf{F} = \mathsf{G}_\lambda(\mathcal{I})$
%
\begin{small}
\begin{align}
\!\! \argmin_{\theta, \phi} \sum_{i=1}^J \sum_{j=1}^{|\mathcal{I}_i|} \sum_{\mathbf{p} \in \mathbf{I}_j} \mathcal{L}(\mathbf{c}_{ij}, \hat{c}(\mathbf{I}_{ij}, \mathbf{p})) + \! \mathcal{L}(d_{ij}, \hat{d}(\mathbf{D}_{ij}, \mathbf{p})) . \label{eq:genNerf}
\end{align}
\end{small}
The key advantage over a vanilla NeRF is that, given a novel scene $\mathcal{I}$, the scene representation function $F$ can be obtained via direct inference without the need of per-scene optimization as in Eq.~\ref{eq:nerf_opt}. We adopt this general formulation for our method. The main drawback is that having to store a dense feature volume as scene representation and having to perform computations on it is memory intensive, hence the resolution limitations of Tab.~\ref{tab:resolution}.

\section{Methodology}

We now present our \sssr{} approach to generalizable NeRFs that requires far less memory and can therefore handle larger resolutions. Our main contribution lies in replacing the scene representation function  $F$ of Eq.~\ref{eq:representation} by one that produces a sparse representation and that we denote as $\mathsf{S} : (\mathbf{x}, \mathbf{r}) \rightarrow \mathbf{f}$. As discussed in the introduction, $\mathsf{S}$ operates in two stages. It first estimates occupancy in a low-resolution----typically $128^3$----volume. The occupied voxels are then supersampled to a higher resolution, typically $512^3$, and the high-dimensional features $\mathbf{f}$ are computed within the resulting high-resolution voxels, as depicted by Fig.~\ref{fig:pipeline}. The sparse representation creates challenges in volume rendering because the ray sampling, querying, and feature aggregations have to be redesigned to handle sparse volumes instead of dense ones, which will also be discussed. 

Note that, in theory at least, instead of adopting the simple occupancy prediction described above, we could have used a more sophisticated octree-like structure, a popular choice to represent 3D geometry sparsely. However, the variable-resolution structure of octrees would have made  volume rendering even more complex for no obvious gain.

In the remainder of this section, we first introduce the representation we use  in both stages for individual voxels and the corresponding features.  We then present the two stages to compute our sparse scene representation $\mathsf{S}$. Finally, we discuss the learning of the $\theta$ and $\phi$ parameters by minimizing Eq.~\ref{eq:genNerf} given such a sparse representation, under a volume rendering framework.

\subsection{Representing Voxels and Features} 
\label{sec:feature_construction}

At low-resolution we use a dense volume while we use a sparse one at high-resolution. Yet, we represent voxels in the same way in both cases.
Given the center point $\mathbf{x}$ of a voxel, we first project it onto the images to obtain $M$-view features
\begin{equation}
\{\mathbf{f}_{I_i}(\pi_i(\mathbf{x}))\}_{i=1}^{M}, 
\label{eq:proj_feature}
\end{equation}
by bilinear interpolation, where $\mathbf{f}_{I_i}$ denotes image features computed by using the Feature Pyramid Network~\citep{Lin17e}. We then write the feature vector associated with $\mathbf{x}$ as 
\begin{align}
    \mathbf{V}(\mathbf{x}) &= \operatorname{MeanVar}(\{\mathbf{f}_{I_i}(\pi_i(\mathbf{x}))\}_{i=1}^{M})  \;  ,
    \label{eq:volume_feature}
\end{align}
where $\operatorname{MeanVar}$ denotes the concatenation of per-channel mean and variance computed from the $M$-view features, and $\mathbf{V}$ denotes the raw dense/sparse feature volumes before a 3D global aggregation process. We take the dimension of $\mathbf{V}(\mathbf{x})$ to be the base feature channel $C_f$.

\subsection{Sparse Scene Representation} 
\label{sec:sparse_representation}

We now describe the two stages---first occupancy prediction and then supersampling the occupied voxels---involved in building a sparse scene representation. 

\parag{Occupancy Prediction}

Assuming that a scene bounding box is given, we voxelize it using a predefined voxel resolution. Due to memory constraints, the resolution cannot be too high. We compute the feature vector using Eq.~\ref{eq:volume_feature} for each voxel, resulting in a dense feature volume $\mathbf{V}_d \in \mathbb{R}^{C \times K^3}$, where $K$ denotes the resolution. We then use a 3D U-Net~\citep{Ronneberger15} $\Psi$ followed by a linear regression head $\mathbf{U}$ to go from raw features $\mathbf{V}_d$ to the occupancy prediction $\mathbf{O} \in \mathbb{R}^{K^3}$. We write 
\begin{align}
    \mathbf{O} &= \mathbf{U}(\Psi(\mathbf{V}_d)) \;
\end{align}

To train  $\Psi$ and $\mathbf{U}$, we rely on the ground-truth surfaces to estimate ground-truth occupancies $\mathbf{O}^{gt}$. In practice, we generate point clouds by merging the ground truth depth maps of each input view. This works better than using the ground-truth point cloud of the scene, which may contain points that are occluded in the input views and thereby impossible to predict. Since the overwhelming majority of voxels are empty in 3D space, we minimize a focal loss~\citep{Lin17f} with focusing parameter $\gamma=2$ during training to counteract the large class-imbalance. We take it to be
\begin{equation}
    \mathcal{L}_{fc} = -\sum_{i,j,k} (1 - p_{ijk})^{\gamma} \log(p_{ijk}) \;
\end{equation}
where
\begin{equation*}
    p_{ijk} = 
    \begin{cases} 
      \mathbf{O}_{ijk} & \text{if } \hat{\mathbf{O}}_{ijk} = 1 \\ 
      1 - \mathbf{O}_{ijk} & \text{if } \hat{\mathbf{O}}_{ijk} = 0
   \end{cases} \;
\end{equation*}
At inference time, we simply threshold the occupancy predictions and apply a dilation process to yield the final binary predictions $\widetilde{\mathbf{O}} = \operatorname{dilate} (\mathbb{I}(\mathbf{O} \geq \tau))$. The predicted occupancy is critical to the performance of our method. We prioritize recall in our method by using a small $\tau$ and employing a dilation process to ensure minimal geometry loss. See our supplementary material for more details.

\parag{Supersampling.} 

The output $\widetilde{\mathbf{O}} \in \mathbb{R}^{K^3}$ is a low-resolution occupancy field. For scene representation purposes, we {\it supersample} the occupied voxels by increasing the resolution of them while ignoring the empty ones. Specifically, for a given voxel, the supersampling operation is performed by placing $s \times s \times s$ regular-grid samples within it, lifting the resolution from $K^3$ to $(sK)^3$. After supersampling, each voxel will be divided into mini-voxels, from which we follow Eq.~\ref{eq:volume_feature} to construct volumetric representations, resulting in raw 3D sparse feature volumes $\mathbf{V}_s \in \mathbb{R}^{N \times C \times s^3}$, where $N$ denotes the number of occupied voxels. We denote each $s \times s \times s$ block as a mini-volume.

We use the SparseUNet implementation from the sparse convolution library \texttt{torchsparse}~\citep{Tang23} to perform 3D feature aggregation, as shown in the middle of the bottom row of Fig.~\ref{fig:pipeline}. The sparse scene representation obtained is 
$
    \mathbf{S} = \Psi^{sp}(\mathbf{V}_s) \; ,
$
where $\mathbf{S} \in \mathbb{R}^{N \times C \times s^3}$ and $\Psi^{sp}$ denotes the 3D SparseUNet.

\subsection{Sparse Volumetric Reconstruction}
\label{sec:sparse_recon}

Minimizing the losses of Eq.~\ref{eq:genNerf}, requires repeatedly evaluating $R$ given sparse scene representation $\mathbf{S}$, which means that $R$ must efficiently handle the sparse nature of our representation. This means that the mechanisms for ray sampling, querying volume locations, and volume rendering used by $R$ have to be redesigned to handle sparse volumes instead of dense ones.

\parag{Ray Sampling.} 

Sampling along the ray has to be adapted because the only meaningful samples are those within occupied voxels. For any given ray, we detect its intersections with every occupied voxel and confine the sampling to ray fragments within occupied ones. We take an arbitrary sampled ray to be $\mathbf{r}(t_i) = \mathbf{o} + t_i \mathbf{d}, \, i = 1, 2, ..., N_s$, where $t_i$ denotes the samples, $\mathbf{o}$ and $\mathbf{d}$ denote ray origin and ray direction respectively.

\parag{Querying $\mathbf{S}$.} 

The volume rendering process is predicated on the ability to query at arbitrary continuous locations along 3D rays, which is easy to achieve in a dense volume but not in sparse ones due to their irregularity. Here we propose a simple yet effective algorithm to query the sparse feature volumes efficiently as shown in Alg.~\ref{alg:query}. The key to making our query algorithm very efficient is our specific data structure: our sparse volumes are a collection of small regular volumes spanning the occupied voxels. We pre-compute a dense lookup table $\mathbf{H}$ to encode the order of stored small volumes in memory and then perform the operations below. 
Please refer to our supplementary material for more details.

\begin{algorithm}
    \caption{Querying Sparse Volumes}
    \label{alg:query}
    \begin{algorithmic}[1]
    \State \textbf{Input:} Querying point $\mathbf{x}$, lookup table $\mathbf{H} \in K^3$, sparse feature volumes $\mathbf{S} \in \mathbb{R}^{N \times s^3 \times C}$.
    \State Find corner points surrounding $\mathbf{x}$, denote their coordinates $\{\mathbf{v}_g^i\}_{i=1}^8$ at scale $(sK)^3$.
    \State Convert $\{\mathbf{v}_g^i\}_{i=1}^8$ at scale $(sK)^3$ to $\{\mathbf{v}_o^i\}_{i=1}^8$ at scale $K^3$ plus local shift $\{\mathbf{v}_l^i\}_{i=1}^8$ at scale $s^3$.
    \State Get corner features from memory by $\mathbf{S}[\mathbf{H}[\mathbf{v}_o^i], \mathbf{v}_l^i, :]$.
    \State Perform trilinear interpolation with corner features.
    \State \textbf{Output:} Interpolated feature $\mathbf{f}_{vol}$.
    \end{algorithmic}
\end{algorithm}

\begin{figure*}[t]
    \centering
    \includegraphics[width=0.99\textwidth]{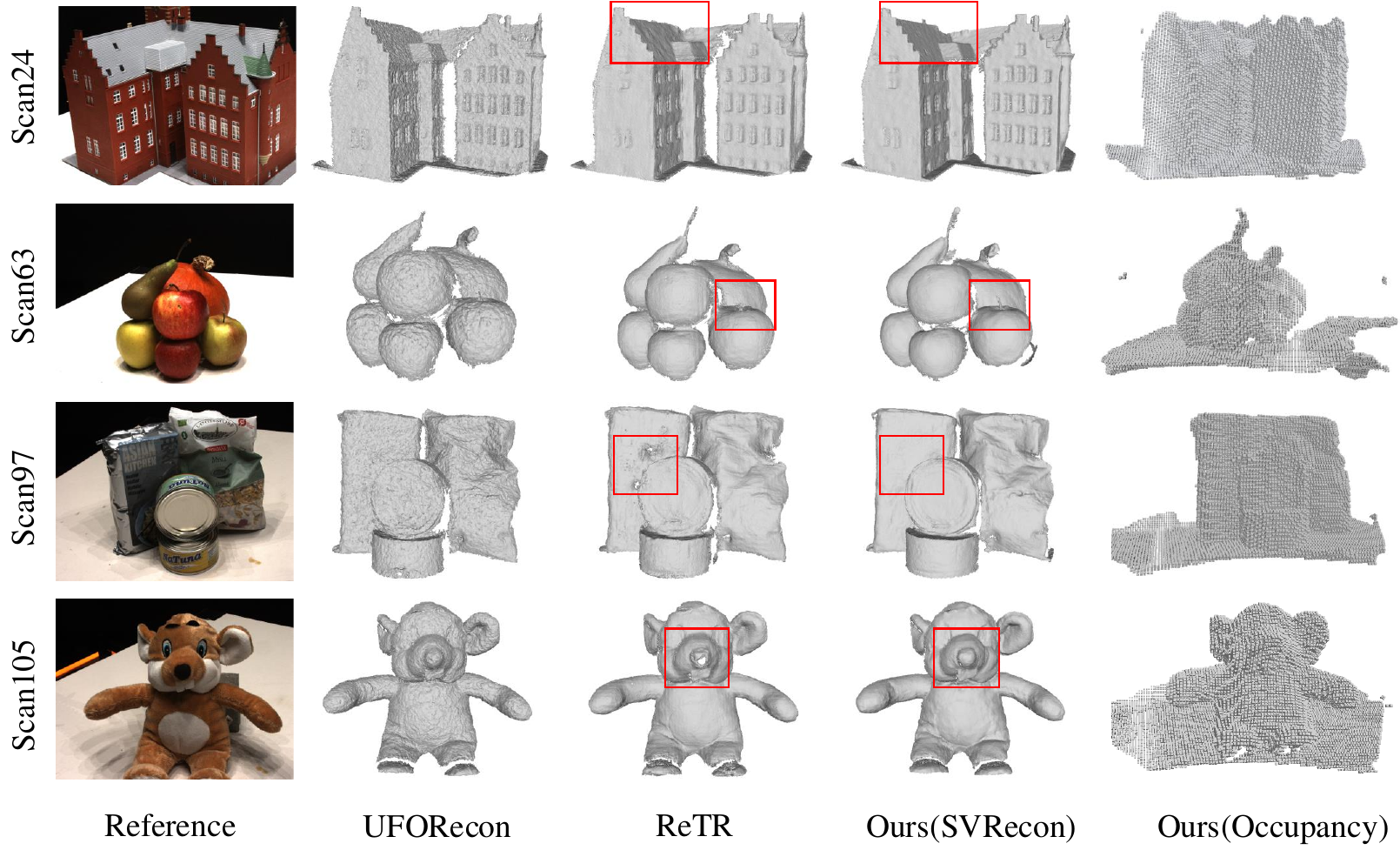}
    \caption{\textbf{Sparse-view surface reconstructions on \texttt{DTU} test scenes.} In the middle columns, we show shaded surfaces for UFORecon, ReTR, and our approach. In the rightmost column, we show the occupancies predicted by the first stage of our method at resolution $128^3$. Voxels are shrunk slightly for visualization purposes. The red rectangles highlight the region with visible differences. }
    \label{fig:qualitative}
\end{figure*}

\parag{Rendering Function $\mathsf{R}$.} 

We adopt the generalized volume rendering equation of~\citep{Liang24} to model $\mathsf{R}$. Specifically, the augmented feature representation for a ray sample $t_i$ is $\mathbf{f}^{r}_i = \text{cat}(\mathbf{f}_{vol},\mathbf{f}_{proj}, \beta)$, where $\beta$ denotes positional encoding~\citep{Wang21i}, $\text{cat}(\cdot)$ denotes concatenation, $\mathbf{f}_{vol}$ denotes queried feature from $\mathbf{S}$, and $\mathbf{f}_{proj}$ denotes the aggregated projection features of Eq.~\ref{eq:proj_feature} using a Transformer network~\citep{Ren23a}. We write
\begin{align} 
    \label{eqn:volume_render}
\mathbf{c}(\mathbf{r}) &= \mathcal{C}\left(\sum_{i=1}^{N_s}\sigma\left( \frac{q(\mathbf{f}^{tok})k(\mathbf{f}_{i}^{occ})^\top}{\sqrt{C}}\right)v(\mathbf{f}_{i}^{r})\right) \\
d\left(\mathbf{r}\right) &= \sum_{i=1}^{N} \sigma\left( \frac{q(\mathbf{f}^{tok})k(\mathbf{f}^{r}_i)^\top}{\sqrt{C}}\right)t_i,
\end{align} 
where $\sigma$ denotes the \textit{softmax} operation, $\mathbf{f}^{tok}$ is a learnable token, $\mathbf{f}_{i}^{occ}$ denotes occlusion-aware feature derived from $\mathbf{f}^r_i$, $q(\cdot),k(\cdot),v(\cdot)$ are linear layers, $C$ denotes feature dimension, and $\mathcal{C}(\cdot)$ is an MLP. To learn the network weights, we minimize the loss function 
\begin{align}
&\mathcal{L} = \mathcal{L}_{\text {color}} + \alpha \mathcal{L}_{\text {depth}}  \\
&=  \frac{1}{N_s} \sum_{i=1}^{N_s} \left\|\mathbf{c}\left( \mathbf{r} \right)-\mathbf{c}_{g}\left( \mathbf{r} \right)\right\|_2 + \frac{\alpha}{N_d} \sum_{i=1}^{N_d} |d\left( \mathbf{r} \right)-d_{g}\left( \mathbf{r} \right)| , \nonumber
\end{align}
where $\mathbf{c}_{g}\left( \mathbf{r} \right)$ and $d_{g}\left( \mathbf{r} \right)$ are ground truth color and depth, respectively, $\alpha$ denotes a weight coefficient to balance the two terms, $N_s$ the number of sampled rays and $N_d$ the number of rays with valid depth.
% !TEX root = ../main.tex
% !TEX spellcheck = en-US

\section{Experimental Results}

\label{sec:experiment}

In this section, we begin by outlining our experimental setup, including datasets and implementation details. Next, we assess our approach both qualitatively and quantitatively on generalizable surface reconstruction and occupancy prediction tasks. We then evaluate the generalization ability of our method on new datasets. Finally, we conduct analyses for different elements in our method and discuss the limitations.

\parag{Datasets.} 

As in previous works~\citep{Liang24,Ren23a,Xu23f,Na24}, we use the \texttt{DTU}~\citep{Aanaes16} dataset for training and main evaluation. It consists of high-resolution images of 124 different scenes under 7 lighting conditions captured under controlled laboratory conditions, each accompanied by accurate camera matrix and laser-scanned ground truth depth map. We use the same evaluation protocol as in earlier work and 3 views as input for each one of the 15 test scenes. In addition to \texttt{DTU}, we adopt more datasets to evaluate the generality of our approach, including \texttt{BlendedMVS}~\citep{Yao20b}, a large-scale synthetic dataset designed for learning-based multi-view stereo, providing rendered images with dense depth maps and camera parameters from diverse 3D scenes; and \texttt{Co3D}~\citep{Reizenstein21}, a large real-world dataset for category-level 3D understanding containing multi-view videos of object instances across many common categories.

\paragraph{Implementation Details.} We use $M = 4$ input views with resolution $640 \times 512$ in training, and $M = 3$ input views with resolution $800 \times 600$ in testing, consistent with previous works. In volumetric feature construction Sec.~\ref{sec:feature_construction}, we use $C=32$ feature channels. Our model is trained for 16 epochs using the Adam optimizer~\citep{Kingma14}; the learning rate is set to $10^{-4}$. 
Please refer to our supplementary material for more details.

% !TEX root = ../main.tex
% !TEX spellcheck = en-US

\begin{table*}[t!]
    \centering
    \small
    \resizebox{1.0\textwidth}{!}{
    \begin{tabular}{lccccccccccccccccc}
    \hline
    Scan & Mean (CD) $\downarrow$ & 24 & 37 & 40 & 55 & 63 & 65 & 69 & 83 & 97 & 105 & 106 & 110 & 114 & 118 & 122  \\
    \hline
    COLMAP~\cite{Schonberger16} & 1.52 & 0.90 & 2.89 & 1.63 & 1.08 & 2.18 & 1.94 & 1.61 & 1.30 & 2.34 & 1.28 & 1.10 & 1.42 & 0.76 & 1.17 & 1.14  \\
    TransMVSNet~\cite{Ding22} & 1.35 & 1.07 & 3.14 & 2.39 & 1.30 & 1.35 & 1.61 & \underline{0.73} & 1.60 & 1.15 & 0.94 & 1.34 & \textbf{0.46} & 0.60 & 1.20 & 1.46  \\
    \hline
    VolSDF~\citep{Yariv21} & 3.41 & 4.03 & 4.21 & 6.12 & 1.63 & 3.24 & 2.73 & 2.84 & 1.63 & 5.14 & 3.09 & 2.08 & 4.81 & 0.60 & 3.51 & 2.18  \\
    NeuS~\citep{Wang21f} & 4.00 & 4.57 & 4.49 & 3.97 & 4.32 & 4.63 & 1.95 & 4.68 & 3.83 & 4.15 & 2.50 & 1.52 & 6.47 & 1.26 & 5.57 & 6.11 \\
    SparseNeuS-ft~\citep{Long22b} & 1.27 & 1.29 & 2.27 & 1.57 & 0.88 & 1.61 & 1.86 & 1.06 & 1.27 & 1.42 & 1.07 & 0.99 & 0.87 & 0.54 & 1.15 & 1.18  \\
    SparseCraft~\cite{Younes24} & \underline{1.04} & 1.17 & \textbf{1.74} & 1.80 & \textbf{0.70} & 1.19 & 1.53 & 0.83 & \textbf{1.05} & 1.42 & 0.78 & \underline{0.80} & 0.56 & \textbf{0.44} & \textbf{0.77} & \textbf{0.84}  \\
    \hline
    PixelNeRF~\cite{Yu21c} & 6.18 & 5.13 & 8.07 & 5.85 & 4.40  & 7.11 & 4.64 & 5.68 & 6.76 & 9.05 & 6.11 & 3.95 & 5.92 & 6.26 & 6.89 & 6.93 \\
    IBRNet~\cite{Wang21i} & 2.32 & 2.29 & 3.70  & 2.66 & 1.83 & 3.02 & 2.83 & 1.77 & 2.28 & 2.73 & 1.96 & 1.87 & 2.13 & 1.58 & 2.05 & 2.09 \\
    MVSNeRF~\cite{Chen21f} & 2.09 & 1.96 & 3.27 & 2.54 & 1.93 & 2.57 & 2.71 & 1.82 & 1.72 & 2.29 & 1.75 & 1.72 & 1.47 & 1.29 & 2.09 & 2.26 \\
    \hline
    SparseNeuS~\cite{Long22b} & 1.96 & 2.17 & 3.29 & 2.74 & 1.67 & 2.69 & 2.42 & 1.58 & 1.86 & 1.94 & 1.35 & 1.50  & 1.45 & 0.98 & 1.86 & 1.87 \\
    VolRecon~\cite{Ren23a} & 1.38 & 1.20 & 2.59 & 1.56 & 1.08 & 1.43 & 1.92 & 1.11 & 1.48 & 1.42 & 1.05 & 1.19 & 1.38 & 0.74 & 1.23 & 1.27  \\
    ReTR~\citep{Liang24} & 1.17 & 1.05 & 2.31 & 1.50 & 0.96 & 1.20 & 1.54 & 0.89 & 1.34 & 1.30 & 0.87 & 1.06 & 0.77 & 0.59 & 1.06 & 1.11  \\
    C2F2NeuS~\citep{Xu23f} & 1.11 & 1.12 & 2.42 & \underline{1.40} & \underline{0.75} & 1.41 & 1.77 & 0.85 & 1.16 & 1.26 & \underline{0.76} & 0.91 & 0.60 & \underline{0.46} & 0.88 & \underline{0.92}  \\
    SuRF~\citep{Peng24b} & 1.05 & 0.85 & 2.35 & 1.48 & 0.88 & 1.17 & \underline{1.39} & 0.74 & \underline{1.14} & 1.24 & 0.84 & \textbf{0.77} & \underline{0.51} & 0.51 & 0.86 & 1.03 \\
    UFORecon~\citep{Na24} & \textbf{1.00} & \underline{0.79} & 2.03 & \textbf{1.33} & 0.87 & \textbf{1.11} & \textbf{1.19} & 0.74 & 1.22 & \underline{1.14} & \textbf{0.71} & 0.89 & 0.59 & 0.56 & 0.90 & 1.02  \\
    \sssr{} & \textbf{1.00} & \textbf{0.72} & \underline{1.98} & 1.44 & 0.83 & \underline{1.12} & 1.40 & \textbf{0.72} & 1.27 & \textbf{1.06} & \underline{0.76} & 0.94 & 0.56 & \textbf{0.44} & \underline{0.87} & 0.93  \\
    \hline
    \end{tabular}
    }
    \vspace{0.2cm}
    \caption{\textbf{Quantitative evaluation results of sparse-view reconstructions on 15 testing scenes of \texttt{DTU} dataset using \textit{Chamfer Distances}.} We test and report results using released codes for VolRecon, ReTR, and UFORecon, other results are sourced from these papers. From top to bottom, the baseline methods are from different categories: 1) Multi-view Stereo (MVS) methods; 2) neural implicit reconstruction methods (requiring per-scene optimization); 3) generalizable neural rendering methods; 4) generalizable surface reconstruction methods. We use \textbf{bold} to indicate best performance, and \underline{underline} to indicate the second-best. ReTR is the closest baseline and also the one we built our approach upon.}
     \label{table:main_results}
\end{table*}

\begin{table*}[t!]
    \centering
    \small
    \resizebox{1.0\textwidth}{!}
    {\begin{tabular}{lccccccccccccccccc}
    \hline
    Scan & AUC@$15^{\circ}\uparrow$ & 24 & 37 & 40 & 55 & 63 & 65 & 69 & 83 & 97 & 105 & 106 & 110 & 114 & 118 & 122  \\
    \hline
    TransMVSNet~\cite{Ding22} & 9.2 & 9.9 & 9.0 & 13.2 & 12.8 & 3.0 & 7.5 & 11.1 & 5.0 & 10.9 & 7.2 & 14.3 & 6.2 & 9.9 & 10.1 & 7.7  \\
    \hline
    SparseCraft~\cite{Younes24} & 19.1 & 22.7 & 15.6 & 26.2 & 25.9 & 14.5 & 14.8 & 15.9 & 10.1 & 10.4 & 14.7 & \textbf{26.9} & \textbf{20.1} & 25.4 & 21.5 & 22.4  \\
    \hline
    MVSNeRF~\cite{Chen21f} & 11.4 & 10.3 & 10.1 & 16.5 & 18.3 & 9.1 & 12.1 & 12.2 & 8.4 & 7.8 & 9.2 & 14.2 & 4.9 & 15.9 & 10.7 & 11.2  \\
    \hline
    SparseNeuS~\cite{Long22b} & 12.1 & 17.3 & 10.7 & 16.5 & 14.7 & 9.3 & 10.9 & 14.6 & 9.1 & 10.4 & 13.1 & 11.4 & 6.3 & 14.9 & 10.8 & 11.8  \\
    VolRecon~\cite{Ren23a} & 8.4 & 8.3 & 7.0 & 14.2 & 12.1 & 5.7 & 8.2 & 7.8 & 5.0 & 5.8 & 6.6 & 10.8 & 3.8 & 11.9 & 8.5 & 10.5  \\
    ReTR~\cite{Liang24} & 16.3 & 20.0 & 11.5 & 26.3 & 20.5 & 14.7 & 14.8 & 15.7 & 9.4 & 12.7 & 12.1 & 19.9 & 14.4 & 21.8 & 15.2 & 15.4  \\
    SuRF~\cite{Peng24b} & 18.9 & 25.4 & \textbf{16.9} & 26.5 & 25.6 & 17.9 & 14.0 & 16.7 & 11.3 & 14.4 & 13.9 & 23.4 & 15.9 & 22.6 & 21.0 & 20.0  \\
    UFORecon~\cite{Na24} & 11.8 & 14.6 & 8.3 & 14.4 & 15.8 & 8.1 & 11.3 & 11.8 & 6.5 & 7.5 & 9.6 & 16.4 & 10.4 & 15.6 & 13.5 & 13.9  \\
    \sssr{} & \textbf{21.0} & \textbf{26.3} & 14.6 & \textbf{28.8} & \textbf{26.4} & \textbf{18.5} & \textbf{18.1} & \textbf{20.3} & \textbf{11.4} & \textbf{15.3} & \textbf{14.9} & 26.3 & 19.6 & \textbf{29.3} & \textbf{22.2} & \textbf{22.9}  \\
    \hline
    \end{tabular}
    }
    \vspace{0.2cm}
    \caption{\textbf{Quantitative evaluation results of sparse-view reconstructions on 15 testing scenes of \texttt{DTU} dataset using \textit{Normal Consistency}.} We test and report results using released codes from the compared methods. We use \textbf{bold} to indicate best performance.}
    \label{table:main_results2}
\end{table*}

\subsection{Main Evaluation Results}

In this section, we present the main reconstruction quality evaluation on the \texttt{DTU} dataset by introducing the evaluation metrics, and reporting quantitative and qualitative results. We subsequently provide occupancy prediction evaluation results in a similar manner.

\parag{Reconstruction Quality Evaluation.} 

\textit{Chamfer Distance} between predicted surfaces and ground truth point clouds has been extensively used and is a standard metric to evaluate surface reconstruction quality. Unfortunately, it lacks awareness of local geometry and density and as a result, does not accurately quantify proper recovery of fine-scale details and reconstructed surface smoothness. Thus, we also report a \textit{Normal Consistency} metric that accounts for surface quality. To evaluate it, we first extract 3D meshes from predicted depth maps and ground truth depth maps using TSDF fusion~\citep{Curless96} and Marching Cube~\citep{Lorensen87}. We then compute the angular differences between normals at closest vertices in the two meshes, and use Area Under the Curve (AUC) up to $15^{\circ}$ in percentage to measure the overall normal consistency robustly. In our supplementary material, we show the angular differences on a specific example.

The quantitative results are shown in Tabs.~\ref{table:main_results} and \ref{table:main_results2}, and the qualitative ones in Fig.~\ref{fig:qualitative}. In terms of \textit{Chamfer Distances}, our method is on par with UFORecon and they both outperform all other methods. However, as can be clearly seen in Fig.~\ref{fig:qualitative}, the surfaces produced by UFORecon are lower-quality and much rougher than ours. The \textit{Normal Consistency} metric quantifies this. In its terms, our method outperforms all the others and especially UFORecon by a large margin. ReTR is the closest baseline to us in methodology and we clearly do better on all metrics and for all scenes.  This confirms that our method delivers  the most regular and smooth surfaces thanks to our sparse high-resolution volumetric representation.

\parag{Occupancy Prediction Evaluation.}  In essence, occupancy prediction is a classification problem for which \textit{Precision} and \textit{Recall} can be used as evaluation metrics. To quantify the savings of our sparse feature volumes scheme in storage, we also compute a \textit{Space Occupation} metric, taken to be the ratio between number of occupied voxels and number of all voxels. There is in general a trade-off between precision and recall, however in our surface reconstruction problem, we prioritize recall and compromise precision such that surface geometry is preserved as much as possible. In practice, 1) we use a conservative threshold of $0.1$ to determine the occupied voxels from the occupancy prediction from the network $\mathbf{O}$; 2) we then further dilate the occupied voxels using a cubic $3 \times 3 \times 3$ kernel to obtain the final occupancy prediction results.

We report our quantitative results in Tab.~\ref{table:occupancy_prediction}, and provide qualitative ones in Fig.~\ref{fig:qualitative}. We achieve a $96.8 \%$ average \textit{Recall}, ensuring the preservation of surface geometry. The false negatives mostly come from the textureless table in the scenes, which is very hard to reconstruct and does not take part in the standard evaluation. The \textit{Precision} is on average at $23.0 \%$, resulting in \textit{Space Occupation} at $1.89 \%$, $4.2$ times the optimal \textit{Space Occupation} at $0.45 \%$. This demonstrates the strong performance of our occupancy prediction method, which recalls most of the surface by keeping only $1.89 \%$ of the voxels on average. 

\subsection{Generalization Ability}

The \texttt{DTU} dataset consists of only indoor objects with identical camera poses, which limits generalization. To validate the generalization ability of our method in surface reconstruction, we follow previous work~\citep{Cao24} to jointly train on both \texttt{DTU} dataset and \texttt{BlendedMVS}~\citep{Yao20b} dataset. For evaluation, we select $10$ scenes from \texttt{Co3D}~\citep{Reizenstein21} dataset to test the generalization ability of our method. We normalize each scene, compute \textit{Chamfer Distance} from predicted points to ground truth points, and use Area Under the Curve (AUC) at threshold $0.05$ as the robust metric. 

\begin{figure}[h]
    \centering
    \includegraphics[width=0.98\linewidth]{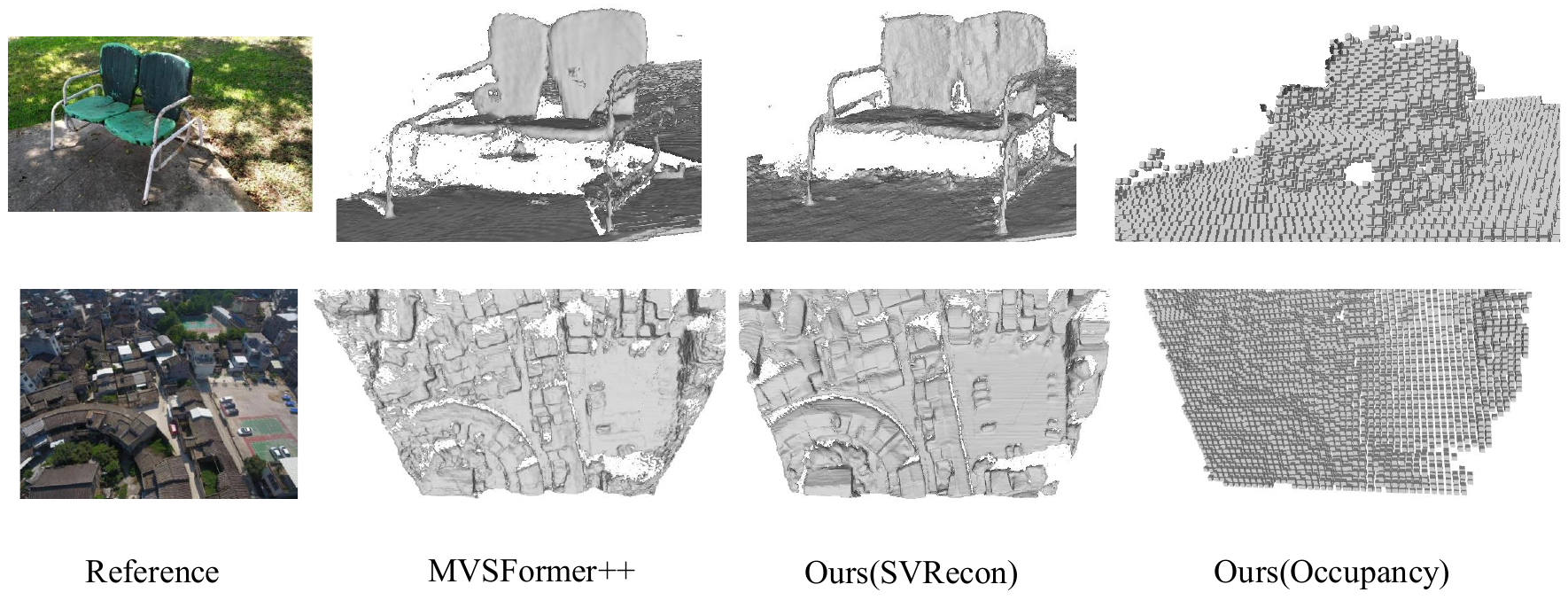}
    \caption{Visualization of cross-dataset generalization results on \textit{bench} scene from \texttt{Co3D} dataset and on a non-object-centric scene from \texttt{BlendedMVS} test set.}
    \label{fig:generalization}
\end{figure}

We use MVSFormer++~\citep{Cao24} as the main competitor, which is trained on the same datasets. We use $M = 3$ input views with resolution $640 \times 480$. The quantitative results are reported in Tab.~\ref{tab:generalization}, and the qualitative results are shown in Fig.~\ref{fig:generalization}, including an example of non-object-centric scenes from \texttt{BlendedMVS} test set. Our method achieves a better performance in reconstruction quality, demonstrating its strong generalization ability.

\begin{table*}[t!]
    \centering
    \resizebox{0.99\textwidth}{!}
    {\fontsize{8}{10}\selectfont
    \begin{tabular}{lccccccccccccccccc}
    \hline
    Scan & Mean & 24 & 37 & 40 & 55 & 63 & 65 & 69 & 83 & 97 & 105 & 106 & 110 & 114 & 118 & 122  \\
    \hline
    Precision & 23.0 & 28.0 & 26.2 & 26.1 & 23.9 & 26.7 & 26.6 & 23.6 & 24.7 & 29.0 & 26.0 & 22.5 & 13.2 & 18.9 & 16.4 & 13.3  \\
    Recall & 96.8 & 98.5 & 91.2 & 90.4 & 98.4 & 92.4 & 94.7 & 99.1 & 97.0 & 98.1 & 97.0 & 97.7 & 98.3 & 99.8 & 99.3 & 99.9   \\
    Space Occupation (Ours) & 1.89 & 2.00 & 2.06 & 1.80 & 1.63 & 2.13 & 1.57 & 1.57 & 2.35 & 2.02 & 2.24 & 1.58 & 1.96 & 1.55 & 1.91 & 2.01  \\
    Space Occupation (GT) & 0.45 & 0.57 & 0.59 & 0.52 & 0.39 & 0.62 & 0.44 & 0.37 & 0.60 & 0.60 & 0.60 & 0.36 & 0.26 & 0.29 & 0.32 & 0.27  \\
    \hline
    \end{tabular}
    }
    \vspace{0.2cm}
    \caption{\textbf{Evaluation results of occupancy predictions on 15 testing scenes of \texttt{DTU} dataset.} We use precision and recall to quantify the performance of occupancy prediction. We also provide space efficiency statistics defined as the ratio between number of occupied voxels and number of all voxels. All reported results are in percentage. }
    \label{table:occupancy_prediction}
\end{table*}

\begin{table*}[t!]
    \centering
    \resizebox{0.99\textwidth}{!}
    {\fontsize{8}{10}\selectfont
    \begin{tabular}{lcccccccccccc}
    \toprule
    Scan & Apple & Backpack & Bench & Bottle & Cake & Couch & Hydrant & Pizza & Bear & Truck & Mean \\
    \midrule
    MVSFormer++~\citep{Cao24} & 0.38 & 0.55 & 0.59 & \textbf{0.38} & \textbf{0.42} & 0.71 & 0.56 & 0.49 & 0.79 & 0.69 & 0.60 \\
    \sssr{} & \textbf{0.54} & \textbf{0.73} & \textbf{0.74} & 0.33 & 0.41 & \textbf{0.88} & \textbf{0.70} & \textbf{0.68} & \textbf{0.92} & \textbf{0.71} & \textbf{0.72} \\
    \bottomrule
    \end{tabular}
    }
    \vspace{0.1cm}
    \caption{\textbf{Evaluation results of cross-dataset generalization on 10 scenes from unseen \texttt{Co3D} dataset.} We use Area Under the Curve (AUC) at threshold $0.05$ as the robust metric. Our method achieves a better performance compared to the baseline, demonstrating its strong generalization ability.}
    \label{tab:generalization}
\end{table*}

\vspace{-1.0em}

\subsection{Further Analyses}

In this section, we conduct further analyses to study the impact of different settings on the final performance of our method. We also provide some novel view synthesis results using our trained volume rendering network. All experiments are conducted on the \texttt{DTU} dataset using the model trained on the same dataset.

\paragraph{Memory Consumption.} We consider only the second stage here, as the first stage of occupancy prediction can be run separately beforehand and does not consume much memory. Practically, our method consumes around 30 GB memory to train with a batch size of $1$, $1024$ sampled rays and $32$ feature channels in the second stage. This is a moderate requirement for modern GPUs, but also prevents us from lifting the resolution even more. In testing, the memory requirement is reduced even more to around 12 GB with the same setting.

\paragraph{Impact of Different Resolutions.} The resolution is of critical importance in the task of surface reconstruction. To verify this point, we use the same occupancy prediction results as in our main experiment and vary the times of supersampling in the second stage of our method. We test our method at $1\times$ and $2\times$ supersampling, leading to resolutions at $128^3$ and $256^3$. Compared to the adopted resolution in our method at $512^3$, the tested resolutions are only lower, because we cannot lift the resolution even more due to memory consumption constraints. The results are presented in Tab.~\ref{tab:ablation_study}. As we can see, there is an abrupt improvement in performance from $128^3$ to $256^3$, and a mild improvement from $256^3$ to $512^3$. Overall, this validates the intuition that the performance will increase as the resolution increases.

\begin{table}[t!]
    \centering
    \small
    \resizebox{0.50\linewidth}{!}{
    \begin{tabular}{c|c}
    \hline
    Settings & Mean (CD) $\downarrow$ \\ \hline
    Resolution @ $128^3$ & 1.27 \\
    Resolution @ $256^3$ & 1.04 \\
    Number of Views @ $5$ & 0.96 \\
    Number of Views @ $4$ & 0.99 \\
    Base Feature Channels @ $16$ & 1.15 \\
    Ours (\sssr) & 1.00 \\ \hline
    \end{tabular}
    }
    \vspace{0.2cm}
    \caption{A study on the impact of different settings on the final performance of our method, including \textbf{resolution}, \textbf{number of input views} and \textbf{number of base feature channels}. }
    \label{tab:ablation_study}
\end{table}

\paragraph{Number of Views.} While our method is trained on $M=4$ input views, it is not restricted to this setting due to the mean and variance feature construction operation as in \cref{sec:feature_construction}. In addition to the $M=3$ setting in our main experiment, we also test the performance of our method with $M=4$ and $M=5$ input views, and the results are given in \cref{tab:ablation_study}. It can be observed that the surface reconstruction quality increases with more input views, as more scene information becomes available.

\paragraph{Number of Base Feature Channels.} The number of feature channels is crucial in determining the effectiveness of scene feature representations. However, more feature channels will also incur more memory burden. In our main experiment, we use $C_f=32$ base feature channels in volumetric feature construction. We also test the performance of our method with $C_f=16$ base feature channels, and the results are provided in \cref{tab:ablation_study}. It shows that reducing the number of base channels will degrade the performance.

\begin{figure}[h]
    \centering
    \includegraphics[width=0.75\textwidth]{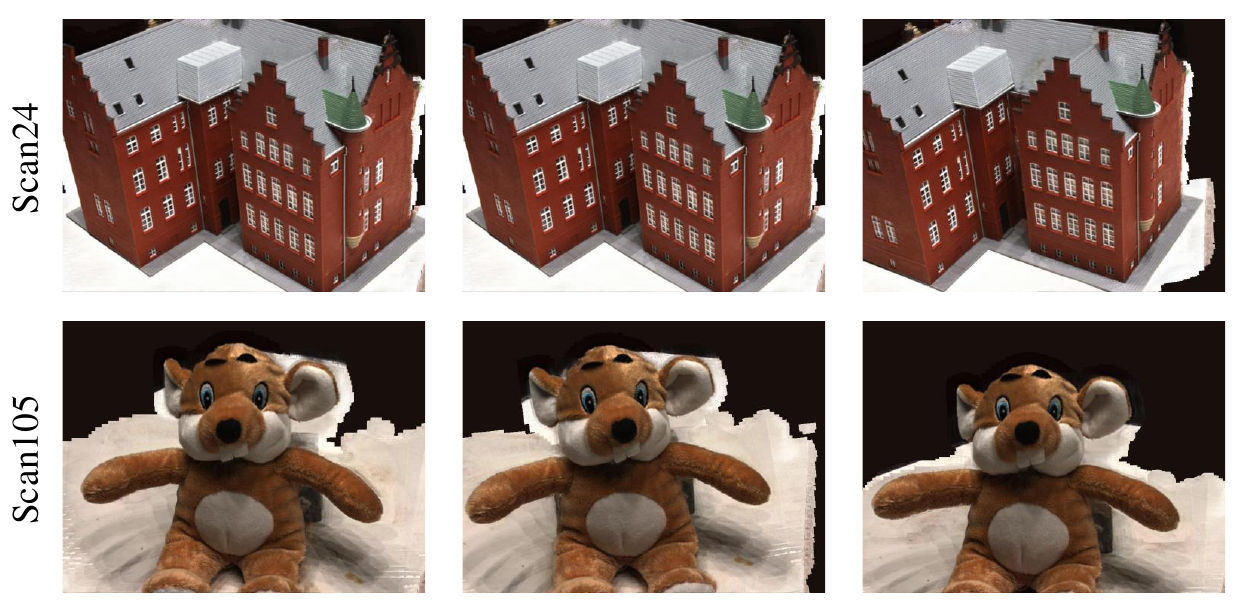}
    \caption{Novel view synthesis examples of our \sssr{} method on scan24 and scan105 from the \texttt{DTU} dataset. The details are preserved well for the foreground object. } 
    \label{fig:novelview}
\end{figure}

\paragraph{Novel View Synthesis.} Our method can also perform novel view synthesis using the trained volume rendering network, and some visual examples are presented in Fig.~\ref{fig:novelview}. Due to our sparse representation, the background is not modeled in the results. 

\subsection{Limitations}

Despite the great performance in high-fidelity reconstruction, our method is limited by the \textbf{accuracy of the occupancy prediction stage}. If the occupancy grid is not accurately predicted, it inevitably leads to incomplete or incorrect surface reconstructions. 

In our experiments, the occupancy prediction network generally works well, but it can struggle in certain cases with large textureless regions or areas with thin structures, resulting in geometry loss. This limitation stems from the inherent challenges of image feature representation for textureless regions and thin structures, which affects 3D predictions in our model as our 3D representation is constructed from 2D image features. We believe that this limitation can be mitigated by incorporating more tailored and powerful image feature representations, which is a promising direction for future work.

In terms of the generalization ability, we have demonstrated that our method can generalize well to unseen datasets in Tab.~\ref{tab:generalization}, even to non-object-centric scenes as shown in Fig.~\ref{fig:generalization}, but it is still limited by the diversity of the training data distribution. For example, if the input views are from distant viewpoints, the performance of our method may degrade. We believe that this limitation can be mitigated by incorporating more diverse training data, following previous work~\citep{Na24}.

\section{Conclusion and Future Work}

In this paper, we propose a two-stage neural surface reconstruction method based on the efficient scene representation of sparse feature volumes. In the first stage, our method is capable of performing accurate occupancy prediction, retaining only around 1.9\% of all voxels as occupied voxels and greatly reducing the memory burden. In the second stage, our method can reconstruct high-quality surfaces by conducting feature-based volume rendering on the constructed sparse feature volumes at a high resolution of $512^3$. Extensive experiments have demonstrated the superiority of our method compared to a variety of existing methods in terms of surface reconstruction quality. In the future, we will explore more efficient schemes that generalize to realistic unbounded scenes with an arbitrary number of views.  

\section*{Acknowledgements}
This work was supported in part by the Swiss National Science Foundation.

\bibliographystyle{splncs04}
\bibliography{main}

@STRING{AAAI     = "AAAI Conference on Artificial Intelligence"}

@STRING{ACM      = "Association for Computing Machinery"}

@STRING{ARXIV    = "arXiv Preprint"}

@STRING{BMVC     = "British Machine Vision Conference"}

@STRING{CVPR     = "Conference on Computer Vision and Pattern Recognition"}

@STRING{ECCV     = "European Conference on Computer Vision"}

@STRING{ICCV     = "International Conference on Computer Vision"}

@STRING{IJCV     = "International Journal of Computer Vision"}

@STRING{JI       = "Journal of Imaging"}

@STRING{MICCAI   = "Conference on Medical Image Computing and Computer Assisted Intervention"}

@STRING{MICRO    = "IEEE/ACM International Symposium on Microarchitecture"}

@STRING{NIPS     = "Advances in Neural Information Processing Systems"}

@STRING{PAMI     = "IEEE Transactions on Pattern Analysis and Machine Intelligence"}

@STRING{SIGGRAPH = "ACM SIGGRAPH"}

@STRING{TOG      = "ACM Transactions on Graphics"}

@article{Aanaes16,
  title = {Large-scale data for multiple-view stereopsis},
  author = {Aan{\ae}s, Henrik and Jensen, Rasmus Ramsb{\o}l and Vogiatzis, George and Tola, Engin and Dahl, Anders Bjorholm},
  journal = IJCV,
  year = {2016}
}

@article{Cao24,
  title = {MVSFormer++: Revealing the Devil in Transformer's Details for Multi-View Stereo},
  author = {Cao, Chenjie and Ren, Xinlin and Fu, Yanwei},
  journal = ARXIV,
  year = {2024}
}

@inproceedings{Chen21f,
  title = {Mvsnerf: Fast generalizable radiance field reconstruction from multi-view stereo},
  author = {Chen, Anpei and Xu, Zexiang and Zhao, Fuqiang and Zhang, Xiaoshuai and Xiang, Fanbo and Yu, Jingyi and Su, Hao},
  booktitle = ICCV,
  year = {2021}
}

@inproceedings{Curless96,
  author = {B. Curless and M. Levoy},
  title = {{A Volumetric Method for Building Complex Models from Range Images}},
  booktitle = {Conference on Computer graphics and Interactive Techniques},
  year = 1996
}

@inproceedings{Ding22,
  title = {Transmvsnet: Global context-aware multi-view stereo network with transformers},
  author = {Ding, Yikang and Yuan, Wentao and Zhu, Qingtian and Zhang, Haotian and Liu, Xiangyue and Wang, Yuanjiang and Liu, Xiao},
  booktitle = CVPR,
  year = {2022}
}

@article{Fua97a,
  author = {P. Fua},
  title = {{From Multiple Stereo Views to Multiple 3D Surfaces}},
  journal = IJCV,
  volume = "24",
  number = "1",
  pages = "19--35",
  month = "August",
  year = 1997
}

@article{Furukawa09b,
  author = {Y. Furukawa and J. Ponce},
  title = {{Accurate, Dense, and Robust Multi-View Stereopsis}},
  journal = PAMI,
  volume = "99",
  year = 2009
}

@inproceedings{Gu20b,
  author = {X. Gu and Z. Fan and S. Zhu and Z. Dai and F. Tan and P. Tan},
  title = {{Cascade Cost Volume for High-Resolution Multi-View Stereo and Stereo Matching}},
  booktitle = CVPR,
  year = 2020
}

@book{Hartley00,
  author = {R. Hartley and A. Zisserman},
  title = {{Multiple View Geometry in Computer Vision}},
  publisher = {Cambridge University Press},
  year = 2000
}

@inproceedings{Huang24b,
  author = {B. Huang and Z. Yu and A. Chen and A. Geiger and S. Gao},
  title = {{2D Gaussian Splatting for Geometrically Accurate Radiance Fields}},
  booktitle = SIGGRAPH,
  year = 2024
}

@inproceedings{Ji17b,
  title = {Surfacenet: An end-to-end 3d neural network for multiview stereopsis},
  author = {Ji, Mengqi and Gall, Juergen and Zheng, Haitian and Liu, Yebin and Fang, Lu},
  booktitle = ICCV,
  year = {2017}
}

@article{Ji20,
  title = {SurfaceNet+: An end-to-end 3D neural network for very sparse multi-view stereopsis},
  author = {Ji, Mengqi and Zhang, Jinzhi and Dai, Qionghai and Fang, Lu},
  journal = PAMI,
  year = {2020}
}

@inproceedings{Kar17,
  author = {A. Kar and C. H{\"a}ne and J. Malik},
  title = {{Learning a Multi-View Stereo Machine}},
  booktitle = NIPS,
  pages = "364--375",
  year = 2017
}

@inproceedings{Kingma14,
  author = {D. P. Kingma and J. Ba},
  title = {{Adam: A Method for Stochastic Optimization}},
  booktitle = ARXIV,
  year = 2014
}

@inproceedings{Kostrikov14,
  author = {I. Kostrikov and J. Gall},
  title = {{Depth Sweep Regression Forests for Estimating 3D Human Pose from Images}},
  booktitle = BMVC,
  year = 2014
}

@article{Kutulakos00,
  author = {K.N. Kutulakos and S.M. Seitz},
  title = {{A Theory of Shape by Space Carving}},
  journal = IJCV,
  volume = "38",
  number = "3",
  pages = "197--216",
  month = "July",
  year = 2000
}

@article{Lhuillier05,
  author = {M. Lhuillier and L. Quan},
  title = {{A Quasi-Dense Approach to Surface Reconstruction from Uncalibrated Images}},
  journal = PAMI,
  year = 2005
}

@inproceedings{Li23c,
  author = {Z. Li and T. M\"uller and A. Evans and R. Taylor and M. Unberath and M. Liu and C. Lin},
  title = {{Neuralangelo: High-Fidelity Neural Surface Reconstruction}},
  booktitle = CVPR,
  year = 2023
}

@inproceedings{Liang24,
  author = {Y. Liang and H. He and Y. Chen},
  title = {{RETR: Modeling Rendering via Transformer for Generalizable Neural Surface Reconstruction}},
  booktitle = NIPS,
  year = 2024
}

@inproceedings{Lin17e,
  author = {T.-Y. Lin and P. Doll{\'a}r and R. Girshick and K. He and B. Hariharan and S. Belongie},
  title = {{Feature Pyramid Networks for Object Detection}},
  booktitle = CVPR,
  year = 2017
}

@inproceedings{Lin17f,
  author = {T.-Y. Lin and P. Goyal and R. Girshick and K. He and P. Doll{\'a}r},
  title = {{Focal Loss for Dense Object Detection}},
  booktitle = ICCV,
  year = 2017
}

@inproceedings{Long22b,
  author = {X. Long and C. Lin and P. Wang and T. Komura and W. Wang},
  title = {{Sparsneus: Fast Generalizable Neural Surface Reconstruction from Sparse Views}},
  booktitle = ECCV,
  year = 2022
}

@inproceedings{Lorensen87,
  author = {W.E. Lorensen and H.E. Cline},
  title = {{Marching Cubes: {A} High Resolution 3{D} Surface Construction Algorithm}},
  booktitle = SIGGRAPH,
  pages = "163--169",
  year = 1987
}

@inproceedings{Mildenhall20,
  author = {Ben Mildenhall and S. P. P. and M. Tancik and J. T. Barron and R. Ramamoorthi and R. Ng},
  title = {{NeRF: Representing Scenes as Neural Radiance Fields for View Synthesis}},
  booktitle = ECCV,
  year = 2020
}

@inproceedings{Na24,
  title = {UFORecon: Generalizable Sparse-View Surface Reconstruction from Arbitrary and Unfavorable Sets},
  author = {Na, Youngju and Kim, Woo Jae and Han, Kyu Beom and Ha, Suhyeon and Yoon, Sung-Eui},
  booktitle = CVPR,
  year = {2024}
}

@inproceedings{Ren23a,
  author = {Y. Ren and F. Wang and T. Zhang and M. Pollefeys and S. S{\"u}sstrunk},
  title = {{Volrecon: Volume Rendering of Signed Ray Distance Functions for Generalizable Multi-View Reconstruction}},
  booktitle = CVPR,
  year = 2023
}

@inproceedings{Ronneberger15,
  author = {O. Ronneberger and P. Fischer and T. Brox},
  title = {{{U-Net}: Convolutional Networks for Biomedical Image Segmentation}},
  booktitle = MICCAI,
  pages = "234--241",
  year = 2015
}

@inproceedings{Schonberger16,
  title = {Pixelwise view selection for unstructured multi-view stereo},
  author = {Sch{\"o}nberger, Johannes L and Zheng, Enliang and Frahm, Jan-Michael and Pollefeys, Marc},
  booktitle = ECCV,
  year = {2016}
}

@inproceedings{Seitz06,
  author = {S.M. Seitz and B. Curless and J. Diebel and D. Scharstein and R. Szeliski},
  title = {{A Comparison and Evaluation of Multi-View Stereo Reconstruction Algorithms}},
  booktitle = CVPR,
  pages = "519--528",
  year = 2006
}

@inproceedings{Shum00b,
  author = {H. Shum and S. B. Kang},
  title = {{Review of Image-Based Rendering Techniques}},
  booktitle = {Visual Communications and Image Processing},
  pages = "2--13",
  year = 2000
}

@inproceedings{Tang23,
  title = {Torchsparse++: Efficient training and inference framework for sparse convolution on gpus},
  author = {Tang, Haotian and Yang, Shang and Liu, Zhijian and Hong, Ke and Yu, Zhongming and Li, Xiuyu and Dai, Guohao and Wang, Yu and Han, Song},
  booktitle = MICRO,
  year = {2023}
}

@inproceedings{Wang21f,
  author = {P. Wang and L. Liu and Y. Liu and C. Theobalt and T. Komura and W. Wang},
  title = {{Neus: Learning Neural Implicit Surfaces by Volume Rendering for Multi-View Reconstruction}},
  booktitle = NIPS,
  year = 2021
}

@inproceedings{Wang21i,
  title = {Ibrnet: Learning multi-view image-based rendering},
  author = {Wang, Qianqian and Wang, Zhicheng and Genova, Kyle and Srinivasan, Pratul P and Zhou, Howard and Barron, Jonathan T and Martin-Brualla, Ricardo and Snavely, Noah and Funkhouser, Thomas},
  booktitle = CVPR,
  year = {2021}
}

@inproceedings{Wang22f,
  title = {Itermvs: Iterative probability estimation for efficient multi-view stereo},
  author = {Wang, Fangjinhua and Galliani, Silvano and Vogel, Christoph and Pollefeys, Marc},
  booktitle = CVPR,
  year = {2022}
}

@inproceedings{Xu23f,
  author = {L. Xu and T. Guan and Y. Wang and W. Liu and Z. Zeng and J. Wang and W. Yang},
  title = {{C2f2neus: Cascade Cost Frustum Fusion for High Fidelity and Generalizable Neural Surface Reconstruction}},
  booktitle = ICCV,
  year = 2023
}

@inproceedings{Yao18,
  title = {Mvsnet: Depth inference for unstructured multi-view stereo},
  author = {Yao, Yao and Luo, Zixin and Li, Shiwei and Fang, Tian and Quan, Long},
  booktitle = ECCV,
  year = {2018}
}

@inproceedings{Yao19,
  title = {Recurrent mvsnet for high-resolution multi-view stereo depth inference},
  author = {Yao, Yao and Luo, Zixin and Li, Shiwei and Shen, Tianwei and Fang, Tian and Quan, Long},
  booktitle = CVPR,
  year = {2019}
}

@inproceedings{Yao20b,
  title = {{BlendedMVS: A Large-Scale Dataset for Generalized Multi-View Stereo Networks}},
  author = {Yao, Yao and Luo, Zixin and Li, Shiwei and Zhang, Jingyang and Ren, Yufan and Zhou, Lei and Fang, Tian and Quan, Long},
  booktitle = CVPR,
  year = {2020}
}

@inproceedings{Yariv20,
  author = {L. Yariv and Y. Kasten and D. Moran and M. Galun and M. Atzmon and B. Ronen and Y. Lipman},
  title = {{Multiview Neural Surface Reconstruction by Disentangling Geometry and Appearance}},
  booktitle = NIPS,
  year = 2020
}

@inproceedings{Yariv21,
  author = {L. Yariv and J. Gu and Y. Kasten and Y. Lipman},
  title = {{Volume Rendering of Neural Implicit Surfaces}},
  booktitle = NIPS,
  year = 2021
}

@inproceedings{Younes24,
  title = {Sparsecraft: Few-shot neural reconstruction through stereopsis guided geometric linearization},
  author = {Younes, Mae and Ouasfi, Amine and Boukhayma, Adnane},
  booktitle = ECCV,
  year = {2024}
}

@inproceedings{Yu21c,
  title = {pixelnerf: Neural radiance fields from one or few images},
  author = {Yu, Alex and Ye, Vickie and Tancik, Matthew and Kanazawa, Angjoo},
  booktitle = CVPR,
  year = {2021}
}

@article{Yu24,
  author = {Z. Yu and T. Sattler and A. Geiger},
  title = {{Gaussian Opacity Fields: Efficient Adaptive Surface Reconstruction in Unbounded Scenes}},
  journal = TOG,
  year = 2024
}

@inproceedings{Zhang25d,
  title = {{Transplat: Generalizable 3D Gaussian Splatting from Sparse Multi-View Images with Transformers}},
  author = {Zhang, Chuanrui and Zou, Yingshuang and Li, Zhuoling and Yi, Minmin and Wang, Haoqian},
  booktitle = AAAI,
  year = {2025}

}

@article{kerbl23,
  author    = {Bernhard Kerbl and Georgios Kopanas and Thomas Leimk{\"{u}}hler and George Drettakis},
  title     = {3{D} {G}aussian {S}platting for Real-Time Radiance Field Rendering},
  journal   = {ACM Transactions on Graphics},
  volume    = {42},
  number    = {4},
  pages     = {139:1--139:14},
  year      = {2023}
}

@inproceedings{Peng24b,
  title={Surface-Centric Modeling for High-Fidelity Generalizable Neural Surface Reconstruction},
  author={Peng, Rui and Shen, Shihe and Xiong, Kaiqiang and Gao, Huachen and Jiao, Jianbo and Gu, Xiaodong and Wang, Ronggang},
  booktitle=ECCV,
  pages={183--200},
  year={2024},
  organization={Springer}
}

@inproceedings{Yang25,
  title={GSRecon: Efficient Generalizable Gaussian Splatting for Surface Reconstruction from Sparse Views},
  author={Yang, Hang and Hui, Le and Qian, Jianjun and Xie, Jin and Yang, Jian},
  booktitle={Proceedings of the IEEE/CVF International Conference on Computer Vision},
  pages={25346--25356},
  year={2025}
}

@inproceedings{Reizenstein21,
  title={Common objects in 3d: Large-scale learning and evaluation of real-life 3d category reconstruction},
  author={Reizenstein, Jeremy and Shapovalov, Roman and Henzler, Philipp and Sbordone, Luca and Labatut, Patrick and Novotny, David},
  booktitle={Proceedings of the IEEE/CVF international conference on computer vision},
  pages={10901--10911},
  year={2021}
}

\clearpage
\setcounter{page}{1}
\title{Generalizable Neural Reconstruction of High-Fidelity Surfaces via Sparse Volumetric Representations -- Supplementary Material}
\titlerunning{Generalizable Neural Reconstruction via Sparse Volumetric Representations}

\author{Aoxiang Fan\inst{1}\orcidlink{0000-0002-2877-9795} \and
Corentin Dumery\inst{1}\orcidlink{0000-0001-5314-7979} \and
Nicolas Talabot\inst{1}\orcidlink{0000-0003-0952-5906} \and
Ming Xu\inst{1}\orcidlink{0000-0002-6478-0582} \and
Hieu Le\inst{2}\orcidlink{0000-0001-7855-2778} \and
Pascal Fua\inst{1}\orcidlink{0000-0002-6702-9970}}

\authorrunning{A. Fan et al.}

\institute{
CVLab, École Polytechnique Fédérale de Lausanne (EPFL), Switzerland\\
\email{\{aoxiang.fan, corentin.dumery, nicolas.talabot, mingda.xu, pascal.fua\}@epfl.ch}
\and
University of North Carolina at Charlotte, USA\\
\email{hle40@charlotte.edu}
}

\maketitle

\section{Querying Sparse Volumes.}
\label{sec:querying_sparse_volumes}

Hereafter, we will use \textit{coarse-voxel} to indicate a voxel at occupancy prediction resolution $K^3$, and \textit{fine-voxel} to indicate a mini-voxel in each mini-volume effectively at resolution $(sK)^3$. To explain the algorithm, we first define a global grid frame, an occupancy grid frame and a local grid frame.
The global grid frame is a hypothetical one at the resolution $(sK)^3$, corresponding to the densified sparse feature volume and each vertex representing a \textit{fine-voxel} center point. The occupancy grid frame is at the resolution of occupancy prediction $K^3$, each vertex representing a \textit{coarse-voxel} center point. Each occupied voxel spans a mini-volume with a local grid frame at resolution $s^3$, where each vertex represents a \textit{fine-voxel} center point as well. For an arbitrary query location $\mathbf{p}$, we need to get the associated features of its eight adjacent vertices from $\mathsf{S}$ to perform trilinear interpolation to obtain $\mathbf{p}$'s feature representation. We detail this functionality in the following.

Knowing the size of a \textit{coarse-voxel}, it is straightforward to find $\mathbf{p}$'s nearest vertices in global grid frame and compute their global coordinates $\{ v_g \mid v_g \in \{0, 1, ..., sK-1\}^3 \}$. Knowing the supersampling rate $s$, we can easily convert the global coordinates into occupancy coordinates $\{ v_o \mid v_o \in \{0, 1, ..., K-1\}^3 \}$ and local grid coordinates $\{ v_l \mid v_l \in \{0, 1, ..., s-1\}^3 \}$. Now we consider the data structure of sparse feature volumes $\mathsf{S} \in \mathbb{R}^{N \times C \times s^3}$. To query the features associated with vertices, we can directly use local grid coordinates for indexing in the last three dimensions. The problem now reduces to finding the right mini-volume index in the first dimension from occupancy coordinates, for which we propose to use a mapping function $\mathbb{H}: \{0, 1, ..., K-1\}^3 \to \mathbb{Z}^+$ to map $v_o$ to the sought index $n$. For that purpose, we define a regular tensor as the dense lookup table that encodes mapped values in its entries. The tensor $\mathbb{H} \in \mathbb{R}^{K^3}$ is at a low resolution, initialized with values of -1 that point to a dummy feature. Then we simply apply boolean indexing in \texttt{PyTorch} to encode mini-volume indices, i.e. $\mathbf{H}[\mathbf{O}] = \{0, 1, ..., N-1\}$. The boolean indexing is consistent with $\mathsf{S}$ in its creation, such that $n = \mathbf{H}[v_o]$ can be used to index $\mathsf{S}$ in the first dimension. Having the associated features of the eight adjacent vertices, trilinear interpolation can be performed for the query result, which completes the query process. 
Given a 3D query point $\mathbf{p}$ with coordinates $[x_p,y_p,z_p]$ in a sparse scene containing N voxels $\{{v_1, ..., v_N}\}$, we need to determine the index $i \in \{1, ..., N\}$ of the sparse voxel $v_i$ containing $\mathbf{p}$. To this end, we construct a 3D hash table $\mathbf{H}$ such that $\mathbf{H}(x_p,y_p,z_p)=i$, mapping spatial coordinates to their corresponding voxel indices. It can be done by performing boolean indexing on a 3D array with ``False'' values everywhere except for each position of the sparse voxels.

\begin{figure*}[t]
    \centering
    \includegraphics[width=1.00\textwidth]{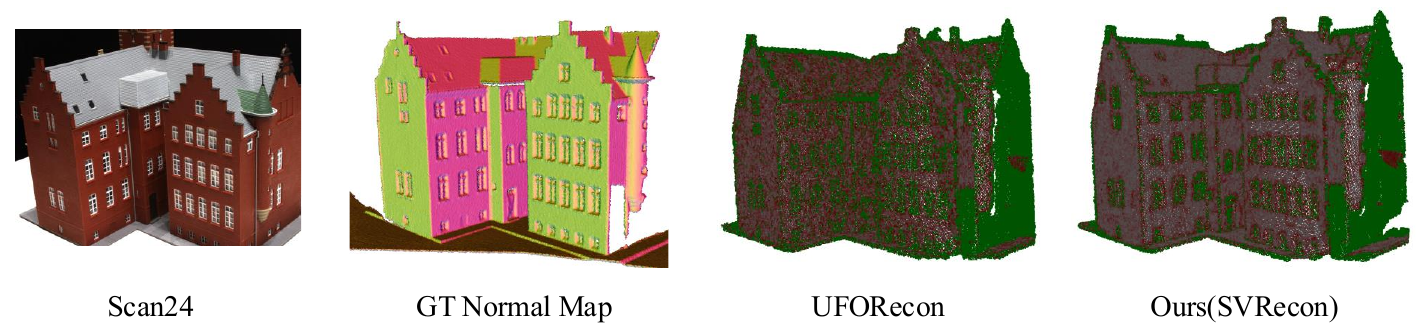}
    \caption{\textbf{\textit{Normal Consistency} evaluation on scan24 from the \texttt{DTU} dataset.} The normal differences are visualized using colors. Errors in the range [$0^{\circ}$, $15^{\circ}$] are color coded linearly from white to red. Error larger than $15^{\circ}$ are shown in green. Our \sssr{} method significantly outperforms UFORecon in \textit{Normal Consistency}.}
    \label{fig:angular}
    \vspace{-6mm}
\end{figure*}

\section{More Implementation Details.}
\label{sec:more_implementation_details}

\paragraph{Method Configurations.} For occupancy prediction, the voxel grid resolution is set to $128^3$. After binarizing the initial predictions using threshold $\tau=0.1$, a morphological dilation process is further applied to the prediction results to maximize the recall of geometry. In the second stage, we supersample each occupied voxel by $s=4$ times in our experiments, leading to resolutions at $512^3$. In ray sampling, we sample $64$ points per-ray both in training and testing. Our model is trained on 4 A100 GPUs for 16 epochs. To reconstruct the surface, we follow the existing works~\citep{Liang24,Ren23a,Xu23f,Na24} to define a virtual rendering viewpoint corresponding to each view by shifting the original camera coordinate frame by $25mm$ along its $x$-axis, and then use TSDF fusion~\citep{Curless96} to merge the rendered depth maps in a volume and extract the mesh from it using Marching Cube~\citep{Lorensen87}.

\paragraph{Network Architectures.} We use Feature Pyramid Network (FPN) for image feature extraction. To obtain projection features from 3D points, we interpolate on the $1/2$ resolution feature map from FPN. For 3D sparse feature volumes we use a sparse UNet architecture, with $4$ encoder layers at dimension $[32,64,128,256]$, bottleneck layer at dimension $256$, and decoder layers at dimension $[256,128,64,32]$. 

\paragraph{Dilation after Network Occupancy Prediction.} To maximize recall, we first use a threshold $\tau=0.1$ to binarize the occupancy predictions from the network. Then we apply a dilation process on the binary occupancy fields. In the dilation process, we apply convolution with a $3 \times 3 \times 3$ kernel on the occupancy fields to compute a score for each voxel, and then set a threshold $3^3*0.1$ to binarize the occupancy fields again as the final occupancy prediction results. By doing this, more voxels near the originally predicted voxels are included to improve recall. The process is written as $\widetilde{\mathbf{O}} = \operatorname{dilate} (\mathbb{I}(\mathbf{O} \geq \tau))$.

\paragraph{Hyperparameter Sensitivity.} We tune two hyperparameters to ensure high recall: the occupancy threshold and the dilation kernel size. We test the impact of different settings of these two hyperparameters on the performance of occupancy prediction in Tab.~\ref{tab:hyperparam_sensitivity}, which shows that our method is not sensitive to the settings of hyperparameters.
\begin{table*}[t]
\centering
\footnotesize
\setlength{\tabcolsep}{4pt}
\begin{tabular}{@{}lccc|ccccc@{}}
\hline
& \multicolumn{3}{c|}{Dilation Kernel Size} 
& \multicolumn{5}{c}{Threshold} \\
\hline
Value 
& 3 & 5 & 7 
& 0.5 & 0.2 & 0.1 & 0.05 & 0.01 \\
Recall (\%) 
& 96.8 & 97.7 & 97.8 
& 92.3 & 95.7 & 96.8 & 97.5 & 98.6 \\
CD 
& 1.00 & 1.00 & 0.99 
& 1.03 & 1.01 & 1.00 & 0.99 & 0.99 \\
\hline
\end{tabular}
\vspace{0.2cm}
\caption{Hyperparameter sensitivity analysis in occupancy prediction.}
\label{tab:hyperparam_sensitivity}
\vspace{-0.2cm}
\end{table*}

\section{More Visualizations.}

\paragraph{Normal Consistency Visualization.} We visualize an example of normal consistency evaluation in Fig.~\ref{fig:angular}.

\vspace{-8pt}

\end{document}